# Fostering User Engagement: Rhetorical Devices for Applause Generation Learnt from TED Talks

**Zhe Liu, Anbang Xu, Mengdi Zhang, Jalal Mahmud and Vibha Sinha**

IBM Almaden Research Center, San Jose, CA 95120 ,USA

{liuzh, anbangxu, zhangmen, jumahmud, vibha.sinha}@us.ibm.com

**Abstract**

One problem that every presenter faces when delivering a public discourse is how to hold the listeners' attentions or to keep them involved. Therefore, many studies in conversation analysis work on this issue and suggest qualitatively constructions that can effectively lead to audience's applause. To investigate these proposals quantitatively, in this study we analyze the transcripts of 2,135 TED Talks, with a particular focus on the rhetorical devices that are used by the presenters for applause elicitation. Through conducting regression analysis, we identify and interpret 24 rhetorical devices as triggers of audience applauding. We further build models that can recognize applause-evoking sentences and conclude this work with potential implications.

## Introduction and Motivation

Many academic studies have been devoted to the tasks of user engagement characterization. However, applause, as the most direct audience reaction, has till now not been fully investigated and understood. The audience do not just clap whenever they like, they do so only at certain points and are continuously looking for appropriate times when applause can possibly occur (Kuo, 2001). Having a deep understanding of audience's applause is important because it will help the speakers to better design their speech with recognizable calls for conjoined response, and to make their presentation more appealing and engageable.

Despite its importance, to date relatively limited work have been conducted on the topic of applause generation, except a few qualitative studies done by social psychologists. Atkinson (1984) first claimed that applause is closely synchronized with a number of actions on the part of the speakers, which he referred to as "rhetorical devices". He identified three rhetorical devices that are effective in evoking applauses, including: contrasts, three-part lists, and projection of names. Heritage and Greatbatch (1986) found five other basic rhetorical devices, including: puzzle-solution, headline-punchline, combination, position taking and pursuit. In addition, new categories were also identified in many recent studies, such as greetings, expressing appreciations, etc. (Bull and Feldman, 2011). To date, research on applause generation has been limited to the analysis of political speeches only. Besides, all of the aforementioned work were conducted using qualitative methods. Many critical questions, such as, What triggers the audience's applause?, When do audience applaud?, etc., remain unanswered.

To address these gaps, this work aims to identify the rhetorical devices for applause elicitation using data-driven methods. To this end, we propose two research questions:

**RQ1:** *What are the rhetorical devices that cause the audiences to applaud during a specific part of a speech?*

**RQ2:** *To what extent the hypothesized rhetorical devices can be used to predict applause generation?*

To answer both questions, we crawl 2,135 TED talk transcripts and conduct quantitative analysis to investigate the factors that could trigger audience's applause. We find that factors such as, gratitude expressions, phonetic structure, projection of names, emotion, etc., have significant effects on applause generation. These identified factors are later used to build machine learning models that can automatically identify applause-evoking segments.

## Method

**Data Collection**

To address our research questions, we write a script to collect 2,135 TED talk videos from 2006 June to 2016 September. We choose TED talks as the dataset of this study because TED talks are widely used as a good source for learning presentation skills across different domains. In addition, for each TED talk there is a spell-checked transcript. Incidences of applause were marked on each transcript in the format of "(Applause)". Relying on the marked applauses, we segment each talk into several chunks, each associate with one incidence of applause. We

---

delete all end-of-speech applause from the transcripts, given that no matter what the speaker says, the audiences will always applause to be polite at the very end. These deletions leave us with 904 talks, comprised of 3,178 times of total applauses.

**Hypothesized Rhetorical Devices**
To answer our RQ1, in this study we propose hypothesized rhetorical devices from 7 different perspectives derived from both theories and observation.

**Linguistic Style** – We rely on LIWC (Pennebaker *et al.*, 2001) to capture language used by speakers to evoke applause. We exclude words from the "Affective or Emotional Processes" sub-category (reasons will be discussed below). We assume the linguistic styles learnt from LIWC (e.g. personal pronouns, achievement, insight) will cover rhetorical devices such as references, and pursuit, etc.

**Emotional Expression** – Speakers can sway the audiences by rousing emotions in their presentations (Petty et al., 2003). We rely on the NRC Emotion Lexicon (Mohammad et al., 2013) for identification of emotion expressions in this study, as it demonstrated more power in emotion detection than LIWC (Brooks et al., 2013). We count the number of words from each of the categories, then normalize them by the total number of words in the sentence.

**Phonetic Structure** – in this work we investigate three phonetic structures: alliteration, rhyme, and homogeneity (Özbal *et al.*, 2013). All three structure are extracted based on the phonetic representation of English words using the CMU pronouncing dictionary (Weide, 1998). Specifically, for alliteration score, we count the number of repeated prefixes appeared in a sentence and normalized by the number of total phonemes. Rhyme score is computed as the ratio of the number of repeated suffix to the total number of phonemes. Rhythmical homogeneity is measured as the ratio of the count of distinct phonemes to the total count of phonemes. We assume that rhetoric devices, such as or slogan-like wordings (e.g. I came, I saw, I conquered) can be well covered by the three phonetic structures.

**Projection of Names** – to test the effectiveness of name-projecting (Atkinson, 1984), we extract all individuals mentioned by matching with names used in Social Security card applications after 1879[1].

**Gratitude Expression** – gratitude expressions are extracted using a self-defined list of words, including grateful\*, gratitud\*, thank\*, appreciate\*, and bless\*, etc.

**Rhetorical Questions** – we simply use question mark as the indicator of questions, as suggested by Li *et al.* (2011).

**Applause-seeking Expression** – Speakers sometimes ask for applause explicitly (e.g. "let's give him an applause"). Applause-seeking expressions are extracted by matching the pattern "applau\*". We use the presence of names, gratitude expressions, rhetorical questions, and applause-seeking expressions as binary features in this work.

**Modeling Approach**
For RQ1, we use logistic regression to examine the associations between the 78 hypothesized rhetorical devices and the evoked applauses. We use the one sentence right before the applause as positive example, since before answering RQ2, we treat the last sentence in front of an applause as its trigger[2]. For negative examples we randomly selected the same number of sentences from the beginning of each applause-segmented chunk. This yields a dataset containing 6,356 records. We conduct a Lasso penalized logistic regression using R glmnet implementation, to control for multicollinearity among predictors. Optimal lambda is determined using 10-fold cross validation.

For RQ2, we build binary classification models to differentiate applause-evoking segments from non-applause-evoking ones using the hypothesized rhetorical devices. We implement the classifiers using different learning algorithms provided in Weka: C4.5, Logistic Regression, Naive Bayes, and SVM. We use the default values in Weka for all learning parameters with 10-fold cross-validation.

So far, all of our analyses are conducted on a single sentence level, this leads us to question if the decision to applaud is only based on what the audience last heard. We wonder if involving more sentences into our applause prediction model would increase its accuracy. To do so, we repeat the classification tasks by increasing the number of sentences before/after an applause. We compare the performances of different models to find the optimal position for the usage of the proposed rhetorical devices.

## Results

**RQ1: Effective Rhetorical Devices Identification**
Table 1 shows the standardized regression coefficient between the hypothesized rhetorical devices and the evoked applause. Here the standardized regression coefficients were calculated through Lasso penalized logistic regression. Significant level for each non-zero coefficient was measured separately and then corrected with False Discovery Rate. Due to the space limit, we only showed the 24 significant rhetorical devices in Table 1.

First, regarding its linguistic styles, we found that sentences with more personal pronouns, either self- (e.g. I, my) or audience-referencing (e.g. you, yours) are more likely to receive applause. Self-referencing can increase the persuasiveness of a speech (Hogan, 2010), whereas audience-referencing can make the listeners feel known and named. Second, applause evoked sentences tend to contain more numbers, certainty (e.g. always, never), and logical expressions (e.g. inclusion and exclusion, such as include, but). All these expressions can make the speakers sound more authoritative and can reinforce his / her statement through voices of expertise (Reyes, 2011). Third, applause-evoking sentences talks more about present than past. Fourth, audiences tend to engage less when the speakers talk about insights (e.g. analyze, conclude) and inhibitions (e.g. barrier, constrain), but applause more when speakers become less formal (when mention e.g. friend, buddy). Additional associations in the usage of fillers (e.g. you know, I mean) and assents (e,g, yes, ok) suggest that speech disfluency negatively affects audience engagement.

---

[1] https://catalog.data.gov/dataset/baby-names-from-social-security-card-applications-national-level-data

[2] We also applied the same approach to varied number of sentences before the applause and the results were consistent. So here we only reported the results based on one sentence.

| Rhetorical Devic- | β | Rhetorical Devic- | β |
|---|---|---|---|
| All pronouns | 0.120 *** | Hearing | -0.013 *** |
| First person singular | 0.043 ** | Social | 0.008 *** |
| Total second person | 0.146 *** | Fillers | -0.007 ** |
| Assents | 0.141 *** | Inclusion | 0.113 *** |
| Prepositions | 0.004 *** | Exclusion | 0.074 *** |
| Numbers | 0.058 *** | Anger | -0.011 *** |
| Insight | -0.065 *** | Alliteration | 0.152 *** |
| Inhibition | -0.010 *** | Rhymes | 0.106 *** |
| Certainty | 0.044 *** | Homogeneity | 0.035 *** |
| Time | 0.051 *** | Name projection | 0.137 ** |
| Past | -0.033 *** | Gratitude | 1.175 *** |
| Present | 0.237 *** | Question | -0.058 *** |

*** $p < 0.001$ ** $p < 0.01$ * $p < 0.05$

*Table 1. Rhetorical devices significantly associated with applause generation and their standardized regression coefficients.*

In terms of the usage of emotional expressions and rhymes, we noticed that applause-evoking sentences tend to express less anger and are more rhythmic. Sentences such as " those are the people I know, those are my friends and family, that is the majority of our society, that is you, that is me" tend to make the whole presentation more powerful and exciting.

Consistent with previous studies on applause generation, our results also indicated that applause will be evoked when the speaker calls for a person's name (e.g. I want to introduce the creators, Alex and Daniel) or expressing thanks or gratitude to the audience or others (e.g. I would like to thank you for listening). However, contrast to previous findings, our results shows a negative association between applause generation and the adaptation of rhetorical questions.

To examine the goodness of fit for the hypothesized rhetorical devices, we further calculated their prediction strength through $R^2$, as well as the correlation coefficient between the predicted values and the true values. Overall, the model with all hypothesized rhetorical devices explained 23% of the variance. Given that there are many other factors, such as speaker's presentation skills, speaking performance, etc., that could also affect the applause generation process, without including any of those determinant, we believed a $R^2$ value of 0.23 for our proposed model is reasonable. The correlation coefficient between the predicted values and the true values is 0.480, which is also quite moderate.

To further understand the phenomenon of applause generation, we next measured the relative importance of each predictor in our model. We calculated the weight for each significant rhetorical device by normalizing the absolute value of β over the sum of all absolute βs (Gilbert and Karahalios, 2009). We plotted the relative importance of each individual rhetorical device in Figure 1. As can be seen from the Figure, the usage of gratitude expressions had the dominant influences in predicting applause generation, followed by the speaker's reference to the present, as well as the adoption of the rhythmic words.

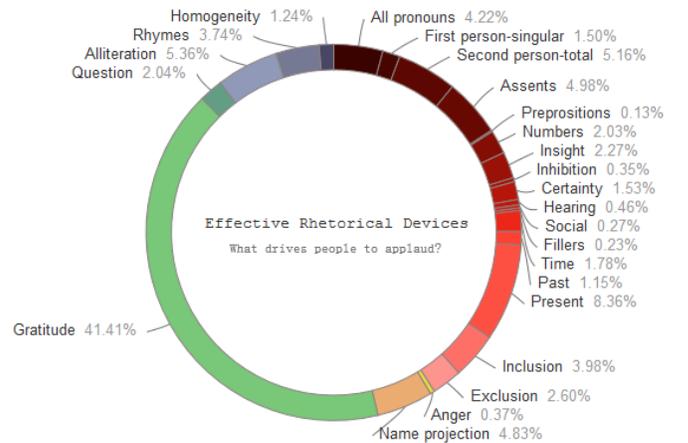

*Figure 1. The relative importance of the 24 significant rhetorical devices in predicting applause generation.*

### RQ2: Applause Prediction

| Rhetorical Devices | Precision | Recall | Accuracy | F1 |
|---|---|---|---|---|
| Linguistic Style | 0.627 | 0.839 | 0.670 | 0.718 |
| Emotion Expression | 0.602 | 0.589 | 0.590 | 0.576 |
| Phonetic Structure | 0.657 | 0.654 | 0.654 | 0.653 |
| Name Projection | 0.510 | 0.542 | 0.510 | 0.576 |
| Gratitude Expression | 0.717 | 0.681 | 0.681 | 0.646 |
| Rhetorical Question | 0.59 | 0.522 | 0.522 | 0.409 |
| Applause Expression | 0.695 | 0.501 | 0.501 | 0.336 |
| Overall | 0.748 | 0.719 | 0.719 | 0.711 |

*Table 2. Classification results obtained on the one sentence basis.*

In Table 2, we showed the classification results for each subset of the rhetorical devices. Due to space limit, we only reported the results from logistic regression as it achieved the best performance. Overall, with all the hypothesized rhetorical devices, our model received a classification accuracy of 71.9%, which is significantly higher than the 50% majority-voted baseline. In terms of each individual feature set, we noticed that consistent with the results of our regression analysis in RQ1, the subset of appreciation and gratitude expressions demonstrated the highest discriminative power. Linguistic styles, emotion expressions, phonetic structure, and applause-seeking expressions are also related to applause elicitation. However, rhetorical features including projection of names and usage of rhetorical questions are not enough by themselves in predicting the incidences of audience applause.

Next, we tested to see if involving more sentences into our prediction model could increase its accuracy. The median number of sentences contained in each applause-segmented chunk is 24. So we decided to try from only one last sentence before the applause to in total six sentences before the applause (last ¼ of the segmented chunk). We plotted in Figure 2 the classification performance of the hypothesized rhetorical devices along the number of sentences selected. As can be seen from Figure 2, the classification accuracy decreased continuously as the number of sentence involved increased. In other words, hearing one sentence is enough to tell if the audience is going to applaud or not. This result also proved the validity of the regression analysis as we conducted in answering RQ1.

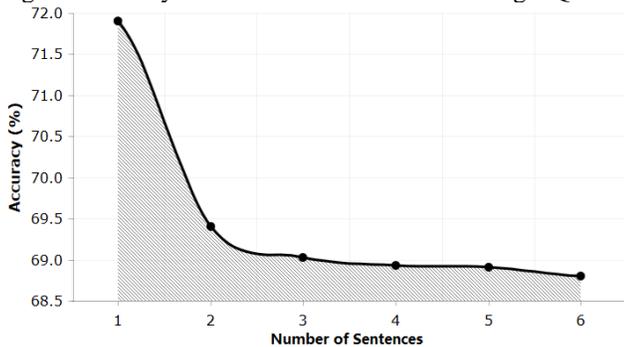

*Figure 2. Classification performance of the hypothesized rhetorical devices along the number of sentences selected.*

## Discussion and Conclusion

This work pays attention to an important but less emphasized social-psychological problem of applause elicitation. We proposed hypothesized rhetorical devices from seven perspectives for applause generation during public discourses. Through regression analysis, we found 24 rhetorical devices significantly associated with the applause generation process and built a model to predict it. We believe that our work is quite unique and important, as we are the first study to understand the rhetorical devices for applause generation at a relatively large scale using quantitative methods. These findings can help system designers to better design tools to assist speakers in generating more engageable speeches. For example, our proposed model can notify the speakers if the audience will applaud as expected at certain point of their speech. Besides, our findings, such as "avoid angry expressions", and "be more prospective, less retrospective", can be used as valuable insights for speakers when preparing their speech. At last, this study is of significant value to social-psychological researchers and can be used as a stepping point for future conversation analysis. A limitation of this study is that it analyzed only English speeches from one platform. However, considering the varied topics covered by TED talks, we believe our results can be also generalized to other platforms. Indeed, for future work, we will evaluate our model on other datasets and other languages for generalizability tests. We are also going to analyze applause with different degrees (only partial audiences applaud) to better understand the effect of the rhetorical devices on applause elicitation under the principle of social proof.